%% file: main.tex
\documentclass[10pt, conference]{IEEEtran}
\usepackage[T1]{fontenc}

\usepackage{amsmath,amssymb,amsfonts}
\usepackage{graphicx}
\usepackage{pifont}
\newcommand{\cmark}{\ding{51}} 
\usepackage{filecontents}

\usepackage{amsthm}
\usepackage{xcolor}
\usepackage[figuresright]{rotating}

\usepackage{graphicx}
\graphicspath{{/}{fig/}}

\usepackage{array}
\usepackage{textcomp}
\usepackage{xcolor}
\usepackage{multirow}
\usepackage{booktabs}

\usepackage{mathtools}
\usepackage{breqn}
\usepackage{float}

\usepackage{pgfplots}
\pgfplotsset{compat=1.18}
\usepgfplotslibrary{groupplots}
\usepackage{pgfplotstable}
\usepackage{svg}

\usepgfplotslibrary{groupplots}
\usepgfplotslibrary{statistics}

\usepackage{caption}
\usepackage{subcaption}
\usepackage{orcidlink}

\usepackage{multirow,tabularx}
\usepackage{hyperref}
\usepackage{flushend}
\usepackage{algorithmic}
\usepackage[vlined, ruled, shortend]{algorithm2e}

\definecolor{AviaColor}{RGB}{228,26,28}
\definecolor{Mid360Color}{RGB}{55,126,184}
\definecolor{OusterColor}{RGB}{77,175,74}

\usepackage{enumitem}

\newlength\figureheight
\newlength\figurewidth
\SetAlCapNameFnt{\footnotesize}
\SetAlCapFnt{\footnotesize}

\title{
    Lidar Variability: A Novel Dataset and Comparative Study of Solid-State and Spinning Lidars\\
}

\author{
    \IEEEauthorblockN{
        \vspace{1em}
        Doumegna Mawuto Koudjo Felix\IEEEauthorrefmark{1}\IEEEauthorrefmark{2}\IEEEauthorrefmark{3},
        Xianjia Yu\IEEEauthorrefmark{1}\IEEEauthorrefmark{3}\orcidlink{0000-0002-9042-3730},
        Jiaqiang Zhang\IEEEauthorrefmark{3}\orcidlink{0000-0002-4509-8115},
        Sier Ha\IEEEauthorrefmark{3}\orcidlink{0009-0000-3617-107X},
        Zhuo Zou\IEEEauthorrefmark{2},
        Tomi Westerlund\IEEEauthorrefmark{3}\orcidlink{0000-0002-1793-2694}
    }
    \IEEEauthorblockA{
        \normalsize
        \IEEEauthorrefmark{1}These authors contributed equally to this work.\\
        \IEEEauthorrefmark{2}School of Information Science and Technology, Fudan Universtiy, China\\
        \IEEEauthorrefmark{3}\href{https://tiers.utu.fi}{Turku Intelligent Embedded and Robotic Systems (TIERS) Lab, University of Turku, Finland}.\\
        Emails: \textsuperscript{1}\{23210720352, zhuo\}@m.fudan.edu.cn, \textsuperscript{2}\{
mawuto.k.doumegna, xianjia.yu, jiaqiang.zhang, sierha, tovewe\}@utu.fi\\[+6pt]
    }
}

\begin{document}

\maketitle
\thispagestyle{empty}
\pagestyle{empty}

\input{sec/00_Abstract.tex}
\IEEEpeerreviewmaketitle

\input{sec/01_Intro}
\input{sec/02_RelatedWorks}
\input{sec/04_Methodology}
\input{sec/05_Experiments}
\input{sec/06_Conclusion}


\section*{Acknowledgment}
This research is supported by the Research Council of Finland's Digital Waters (DIWA) flagship (Grant No. 359247).

\bibliographystyle{unsrt}
\bibliography{bibliography}

\end{document}

%% file: sec/00_Abstract.tex

\begin{abstract}\label{sec:abstract}%
Lidar technology has been widely employed across various applications, such as robot localization in GNSS-denied environments and 3D reconstruction. 
Recent advancements have introduced different lidar types, including cost-effective solid-state lidars such as the Livox Avia and Mid-360. The Mid-360, with its dome-like design, is increasingly used in portable mapping and unmanned aerial vehicle (UAV) applications due to its low cost, compact size, and reliable performance. 
However, the lack of datasets that include dome-shaped lidars, such as the Mid-360, alongside other solid-state and spinning lidars significantly hinders the comparative evaluation of novel approaches across platforms.
Additionally, performance differences between low-cost solid-state and high-end spinning lidars (e.g., Ouster OS series) remain insufficiently examined, particularly without an Inertial Measurement Unit (IMU) in odometry.

To address this gap, we introduce a novel dataset comprising data from multiple lidar types, including the low-cost Livox Avia and the dome-shaped Mid-360, as well as high-end spinning lidars such as the Ouster series. Notably, to the best of our knowledge, no existing dataset comprehensively includes dome-shaped lidars such as Mid-360 alongside both other solid-state and spinning lidars. In addition to the dataset, we provide a benchmark evaluation of state-of-the-art SLAM algorithms applied to this diverse sensor data.
Furthermore, we present a quantitative analysis of point cloud registration techniques, specifically point-to-point, point-to-plane, and hybrid methods, using indoor and outdoor data collected from the included lidar systems. The outcomes of this study establish a foundational reference for future research in SLAM and 3D reconstruction across heterogeneous lidar platforms.

\end{abstract}

\begin{IEEEkeywords}

Point Cloud Matching; Lidars; Dataset; Lidar Odometry; Solid-state lidars; SLAM Benchmarking;

\end{IEEEkeywords}

%% file: sec/01_Intro.tex

\section{Introduction}\label{sec:introduction}
Lidar has become a critical technology in robotics and 3D reconstruction applications. 
It significantly enhances robotic perception when integrated with other sensors, including conventional cameras, event cameras, inertial measurement units (IMUs), and various other sensory devices~\cite{zhang2025event, bai2022faster, wu2022deep}. 
In particular, lidar has emerged as one of the primary sensors for robot localization in GNSS-denied or unreliable environments. In such scenarios, it is commonly used together, significantly improved by, and dependent on IMU~\cite{bai2022faster,he2023point}.

\begin{figure}
    \centering
    \includegraphics[width=0.49\textwidth]{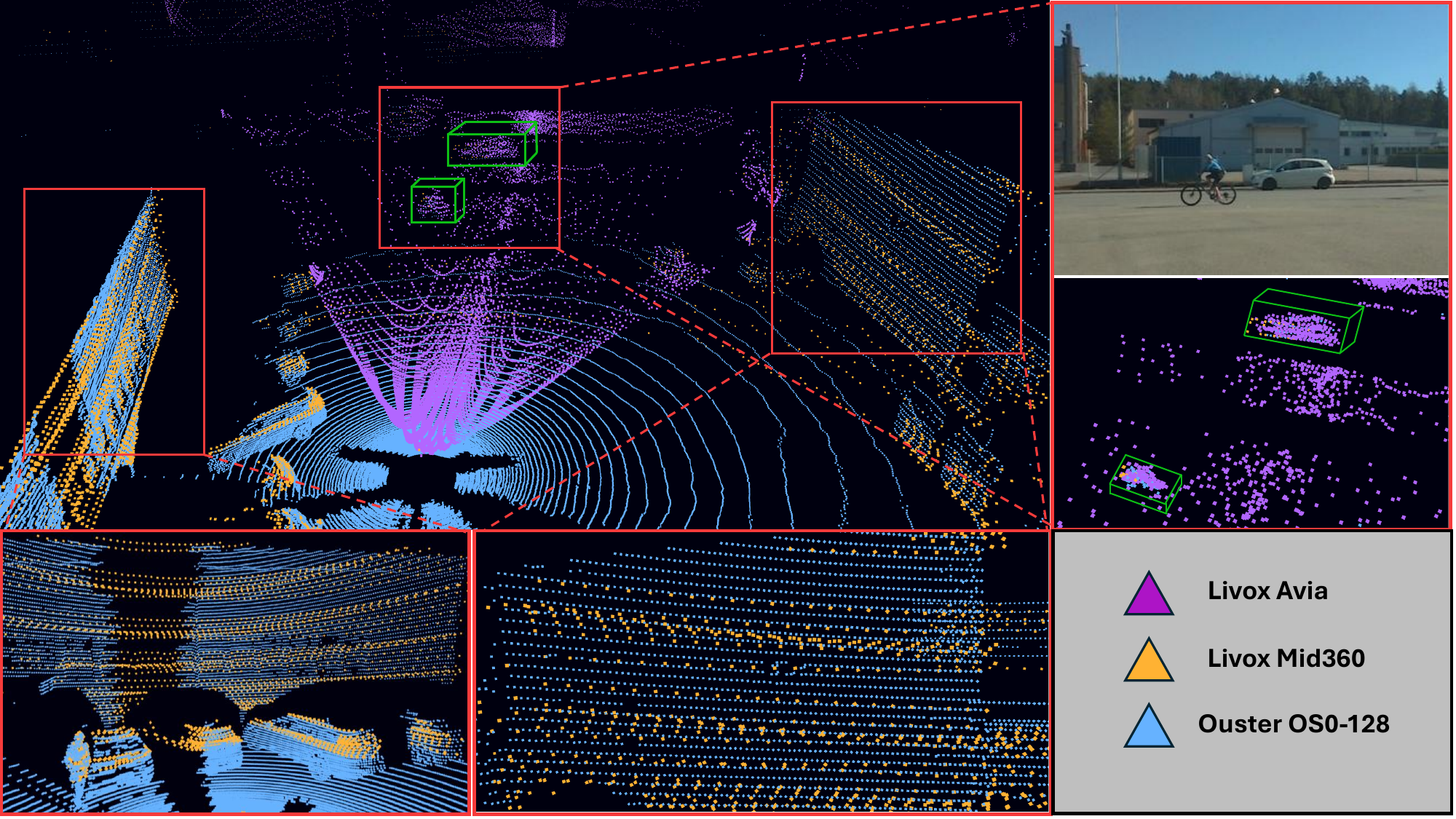}
    \caption{The point cloud differences of the objects under three different types of lidars.}
    \label{fig:concept}
    \vspace{-1em}
\end{figure}

In recent years, lidar technology has evolved considerably for high-end models and more cost-effective alternatives. 
Traditional spinning lidars typically use a rotating mechanism to continuously scan the environment with a laser beam, producing a 360\textdegree\, point cloud over time.
However, solid-state lidars employ no moving parts and instead rely on electronic beam steering or flash illumination to capture depth information. This results in more compact, robust, and energy-efficient designs,albeit often with a narrower field of view (FoV).
Livox lidars have garnered increasing attention among various solid-state lidar technologies due to their unique non-repetitive scanning patterns and cost-effective designs~\cite{lin2020loam, bai2022faster}.
These Livox sensors provide varying fields of view (FoV), such as the Livox Avia, which has a limited FoV, and the Livox Mid-360, which features a dome-shaped design. The latter, in particular, offers a cost-effective yet accurate solution for 3D Lidar-based simultaneous localization and mapping (SLAM). Notably, the Livox Mid-360’s unique design makes it highly suitable for various mapping applications. However, to the best of our knowledge, there is no dataset containing all these lidar types. 

Fig.~\ref{fig:concept} shows the point cloud differences of the same objects under different lidars. Studies have demonstrated that solid-state lidar systems like the Livox Mid-360 deliver competitive performance~\cite{lin2020loam}. Despite these advancements, they typically exhibit a more irregular scanning pattern and sometimes lower resolution than high-end spinning models. These lidar data characteristics (i.e., density, resolution, and scanning pattern) may potentially affect the accuracy of odometry in different types of environments accordingly. However, nowadays, lidar-based odometry is highly dependent on IMUs. It is challenging and lacks independent research on the influence that these lidar features will bring into the point cloud registration or matching, particularly for different types of point cloud registration techniques, such as point-to-point and point-to-plane methods, in the absence of IMU assistance for forward propagation remains unclear. 

To address this gap, this study introduces a novel dataset incorporating all these lidar types, with a motion capture (MoCap) system providing odometry ground truth. Together with the dataset, we provide benchmarking with the current state-of-the-art SLAM methods on the dataset. We then conduct a quantitative analysis to compare the performance of different point cloud matching techniques, including point-to-point, point-to-plane, and hybrid methods. The findings of this study aim to serve as a valuable reference for researchers and practitioners in the field.

The remainder of this document is organized as follows: Section II introduces the recent development of lidars, their datasets, and the state-of-the-art SLAM and point cloud matching methods. Section III then describes this study's methodology. Section VI presents the experimental results, while Section VII concludes the work and outlines directions for future research.

%% file: sec/02_RelatedWorks.tex

\section{Related Work} \label{sec:related_work}
\subsection{Advancement of Lidars}
Lidar technology has seen significant advancements in recent years, leading to the development of high-end spinning Lidar systems and more affordable solid-state options. Traditionally, spinning Lidar sensors, such as those from Velodyne and Ouster, have dominated the market due to their high accuracy and wide field of view (FoV)~\cite{zhang2024streambasedgroundsegmentationrealtime}. These sensors, commonly used in autonomous vehicles, typically operate at fixed frequencies (e.g., 10-20\, Hz) and offer superior scanning capabilities thanks to their rotating laser scanners, which create detailed 3D maps of the environment. Beyond serving as a source for point cloud data, Lidar can also generate images, where pixels are encoded by the intensity of the ambient light or reflected infrared light emitted by itself. This is more evident nowadays when the Lidar resolution increases, such as with Ouster Lidar.  This presents a potential way for applying conventional computer vision methods to these Lidar-generated images~\cite{ha2024enhancing,zhang2023lidar,sier2023uav,yu2023general}.

In contrast, solid-state Lidar systems, such as the Livox Avia and Livox Mid-360, have introduced substantial cost reductions, compactness, and increased adaptability for various applications~\cite{electronics12173633}. The Livox Mid-360 is particularly noteworthy for its unique dome-shaped design, which provides a 360\textdegree\quad horizontal FoV and a vertical range of approximately 25\textdegree. This design enables a more compact solution, making it suitable for portable mapping systems, unmanned aerial vehicles (UAVs), and robotic applications in GNSS-denied environments~\cite{redovnikovic2025affordable}.The Mid-360 is a cost-effective alternative for applications requiring less dense point clouds, such as SLAM (Simultaneous Localization and Mapping) in mobile robots and 3D reconstruction tasks~\cite{electronics12173633}.

Recent advancements in the Lidar industry have made these technologies more accessible, and developments in Livox sensors indicate promising improvements in point cloud density and overall performance. As Lidar systems continue to diversify, the integration of various types of solid-state, spinning, and hybrid systems has paved the way for more comprehensive datasets that can serve as valuable resources for future research~\cite{zhang2024streambasedgroundsegmentationrealtime,redovnikovic2025affordable}.

These advancements also suggest the potential for further optimization in point cloud matching techniques. Research has concentrated on understanding how the specific characteristics of each Lidar type impact the accuracy of 3D reconstructions. Despite technological progress, there is still a lack of datasets that comprehensively combine different Lidar types for benchmarking point cloud matching performance~\cite{electronics12173633}.


\input{tbs/lidar-datasets}

\subsection{Lidar Datasets}
Lidar datasets are crucial for advancing point cloud processing, robotic localization, 3D reconstruction, and Simultaneous Localization and Mapping (SLAM). Over the years, various datasets have been created, primarily for applications in autonomous driving, industrial mapping, and urban modeling. Despite this increasing availability, a significant gap remains, particularly regarding datasets representing various Lidar types encompassing high-end spinning Lidars, low-cost solid-state Lidars, and dome-shaped Lidars such as the Livox Mid-360.

Most existing benchmark datasets predominantly focus on spinning Lidar systems, exemplified by the widely used KITTI dataset~\cite{Geiger2013IJRR} and the Oxford RobotCar dataset~\cite{RobotCarDatasetIJRR}, both incorporating Velodyne HDL-64E spinning Lidars. Similarly, datasets like ApolloScape~\cite{wang2019apolloscape} and Ouster’s research datasets~\cite{ouster} emphasize high-resolution spinning systems. Although these resources have proven invaluable for research on SLAM and autonomous driving, they do not comprehensively address the broader spectrum of Lidar sensor modalities. In particular, datasets combining spinning, solid-state, and dome-shaped Lidars remain rare.

One notable advancement is the TIERS Multi-Modal Lidar Dataset~\cite{li2022dataset}, which includes Velodyne, Ouster OS-series, and Livox Avia sensors to support benchmarking general-purpose localization and mapping algorithms. While the TIERS dataset represents an essential step toward multi-modal sensor integration, it still lacks coverage of dome-shaped sensors such as the Livox Mid-360. This particular sensor’s unique 360° horizontal field of view and approximately 25° vertical span make it especially suitable for portable mapping and UAV applications in GNSS-denied environments~\cite{livox}.

Recently, two additional datasets have emerged that attempt to tackle Lidar-based odometry and registration challenges. The GEODE dataset~\cite{chen2024heterogeneouslidardatasetbenchmarking} was developed to benchmark localization robustness in degenerate environments such as corridors and open spaces. Although GEODE includes heterogeneous Lidar configurations and focuses on structure-scarce conditions, a relevant concern to our study is that it does not incorporate dome-shaped sensors or assess point cloud matching performance in IMU-free settings.

Similarly, the CTE-MLO dataset~\cite{shen2025ctemlocontinuoustimeefficientmultilidar} concentrates on continuous-time multi-Lidar odometry, capturing field data across structured and unstructured environments. Although it integrates spinning and solid-state Lidars, the dataset's organization varies by platform: for instance, Livox Mid-360 is mounted only on a MAV, while ground vehicles primarily utilize Ouster and Robosense sensors. This platform-specific variation limits direct cross-modality comparisons under consistent environmental conditions. Furthermore, CTE-MLO strongly emphasizes Lidar-inertial fusion rather than purely geometry-based point cloud matching, which falls outside the scope of this study.

Despite these recent developments, no publicly available dataset to date comprehensively integrates high-end spinning Lidars, low-cost solid-state Lidars, and dome-shaped Lidars. This limitation restricts researchers’ ability to systematically benchmark point cloud registration methods across modalities, especially when considering differences in scan patterns (regular vs. irregular) and point cloud densities.

To address this gap, this study introduces a novel multi-Lidar dataset that includes Livox Avia, Livox Mid-360, and Ouster OS-series sensors.
It serves as a comprehensive benchmarking resource, enabling detailed evaluations of point cloud matching techniques in IMU-free contexts and facilitating future research into multimodal Lidar integration for SLAM and 3D reconstruction applications.

\subsection{Point Cloud Matching}

Point cloud matching is a core component of 3D reconstruction, SLAM, and robotic navigation. It enables accurate alignment of spatial data from different sensors. Classic approaches include point-to-point, point-to-plane, and hybrid Iterative Closest Point (ICP) algorithms, each tailored to specific geometric and data-quality conditions.

\textbf{Point-to-point ICP} minimizes the Euclidean distance between corresponding points in two point clouds via iterative optimization~\cite{121791}:

\[
\sum_{i=1}^{N} \left\| p_i - T(q_i) \right\|^2
\]

Where \( p_i \) and \( q_i \) are matched points in the source and target clouds, and \( T(q_i) \) is the transformed target. While efficient, this method is sensitive to noise, outliers, and sparsity.

\textbf{Point-to-plane ICP} refines alignment by minimizing the distance from a point in the source to the tangent plane at the corresponding target point~\cite{132043}:

\[
\sum_{i=1}^{N} \left\| p_i - (q_i + \lambda n_i) \right\|^2
\]

Here, \( n_i \) is the surface normal at \( q_i \), and \( \lambda \) adjusts the projection along the normal. This approach improves convergence in structured environments with planar surfaces.

\textbf{Hybrid ICP} combines the two formulations using a weighted error term~\cite{SophiaZhangZhengyou1992}:

\[
\alpha \sum_{i=1}^{N} \left\| p_i - T(q_i) \right\|^2 + (1 - \alpha) \sum_{i=1}^{N} \left\| p_i - (q_i + \lambda n_i) \right\|^2
\]

where \( \alpha \in [0,1] \) balances the influence of point-to-point and point-to-plane terms. This strategy enhances robustness in scenes with mixed geometries or varying point densities.

Each variant presents trade-offs: point-to-point ICP is fast but less tolerant to irregular data; point-to-plane excels in structured settings; hybrid ICP offers flexibility by adapting to scene complexity. In this study, we evaluate the effectiveness of the three ICP variants on multi-modal Lidar datasets, including Livox Mid-360.  The goal is to assess their effectiveness in real-world applications, particularly when dealing with complex, multisource point clouds.


%% file: tbs/lidar-datasets.tex

\begin{table*}[t]
    \centering
    \caption{Comparison of LiDAR Sensors Across Datasets}
    \label{tab:lidar-datasets}
    \resizebox{\textwidth}{!}{%
    \begin{tabular}{llllll}
        \toprule
        \multirow{2}{*}{\textbf{Dataset}} & \multirow{2}{*}{\textbf{LiDAR Sensors Included}} & \multicolumn{3}{c}{\textbf{Type of LiDAR}} & \multirow{2}{*}{\textbf{Focus Area}} \\
        & & Spinning & Solid-State & Dome & \\
        \midrule
        \textbf{KITTI} & Velodyne HDL-64E & \cmark & & & Autonomous Driving, SLAM \\
        \textbf{Oxford RobotCar} & Velodyne HDL-64E, LMS-151, LD-MRS 3D & \cmark & & & Urban Navigation, Long-Term SLAM \\
        \textbf{Ouster (2020)} & OS0, OS1, OS2, OSDome & \cmark & & \cmark & Sensor Benchmarking \\
        \textbf{Livox Simu-dataset} & Livox Horizon, Tele-15 & & \cmark & & Object Detection, SLAM \\
        \textbf{TIERS Lab Dataset} & Velodyne VLP-16, Ouster OS0-128, OS1-64, Livox Horizon, Livox Avia & \cmark & \cmark & & General-Purpose SLAM \& Mapping \\
        \textbf{ApolloScape} & Riegl VMX-1HA (VUX-1HA Laser Scanners), Velodyne HDL-64E & \cmark & & & Urban Mapping, High-Precision 3D Reconstruction \\
        \textbf{GEODE Dataset} & Velodyne VLP-32C, Ouster OS1-32, Livox Avia & \cmark & \cmark & & Localization under Degenerate Conditions \\
        \textbf{CTE-MLO Dataset} & Ouster OS1-64, Robosense Helios 32, Robosense M1, Livox Mid-360 & \cmark & \cmark & \cmark$^\dagger$ & Continuous-Time Multi-LiDAR Odometry \\
        \textbf{Ours (New Dataset)} & Livox Avia, Ouster OS0-128, Livox Mid-360 & \cmark & \cmark & \cmark & LiDAR Variability \& Point Cloud Matching \\
        \bottomrule
    \end{tabular}
    }

    \vspace{0.5em}
    \begin{minipage}{\textwidth}
    \small
    \textbf{Note:} $^\dagger$ \textit{The dome-shaped LiDAR (Livox Mid-360) in the CTE-MLO dataset is mounted only on the MAV platform; it is not integrated alongside spinning or solid-state LiDARs on ground vehicles.}
    \end{minipage}
\end{table*}

%% file: sec/04_Methodology.tex

\section{Methodology}

This section outlines the methodology for processing and analyzing point cloud data collected from multiple Lidar sensors. The study follows a structured workflow that includes data acquisition, preprocessing, point cloud registration utilizing three variants of the Iterative Closest Point (ICP) algorithm, and evaluating trajectory accuracy against ground truth data.

\subsection{Hardware}
\begin{figure}[h]
    \centering
    \includegraphics[width=0.48\textwidth]{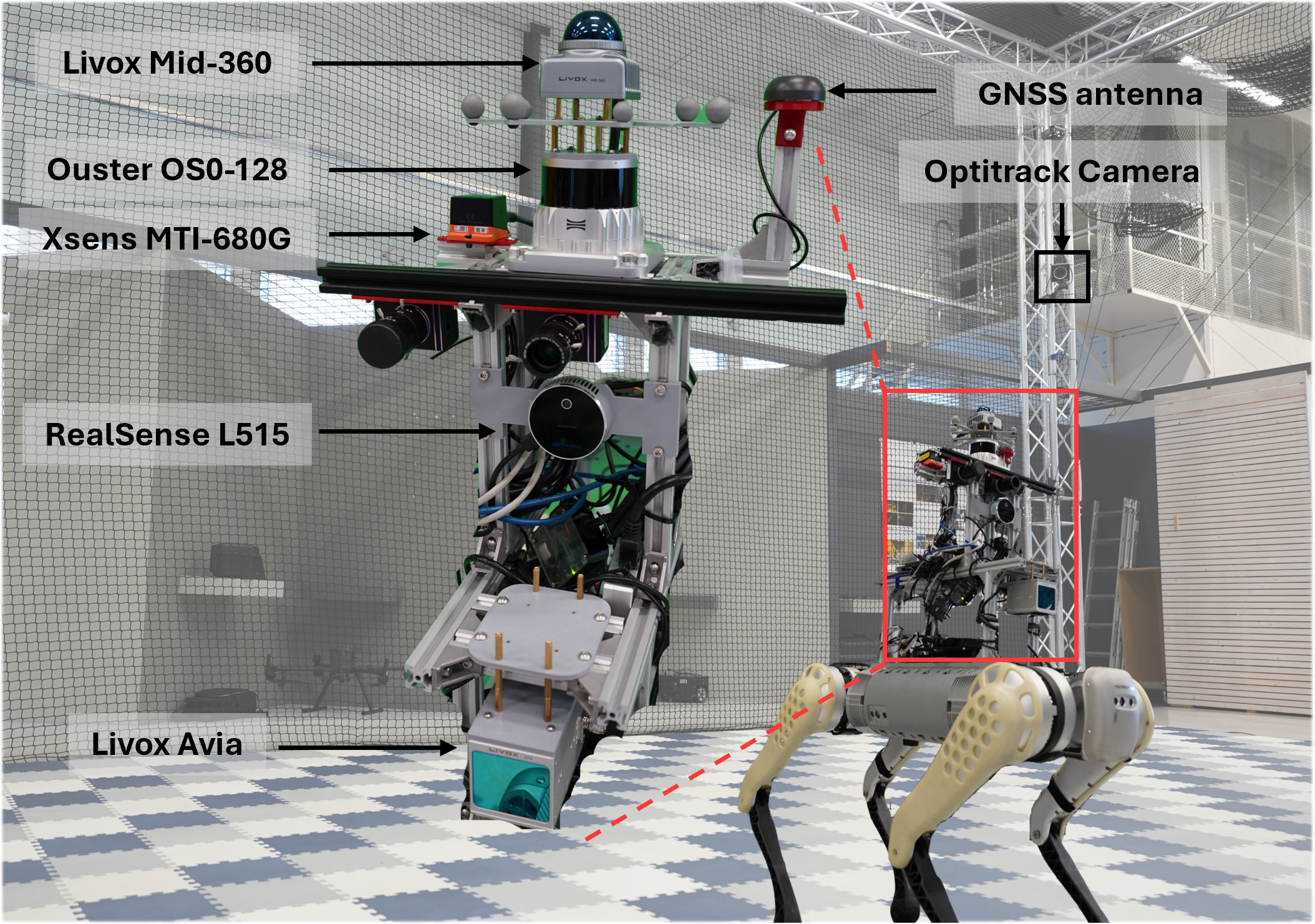}
    \caption{The hardware used for the data collection.}
    \label{fig:hardware}
\end{figure}

The hardware used for the data collection in this study is illustrated in Fig.~\ref{fig:hardware}.  The moving robot platform is a Unitree B1 quadruped robot equipped with various sensors, including Livox Mid-360 and Avia, Ouster OS0-128, RealSense L515, and Xsens MTI-680G.

\begin{figure}
    \centering
    \includegraphics[width=0.48\textwidth]{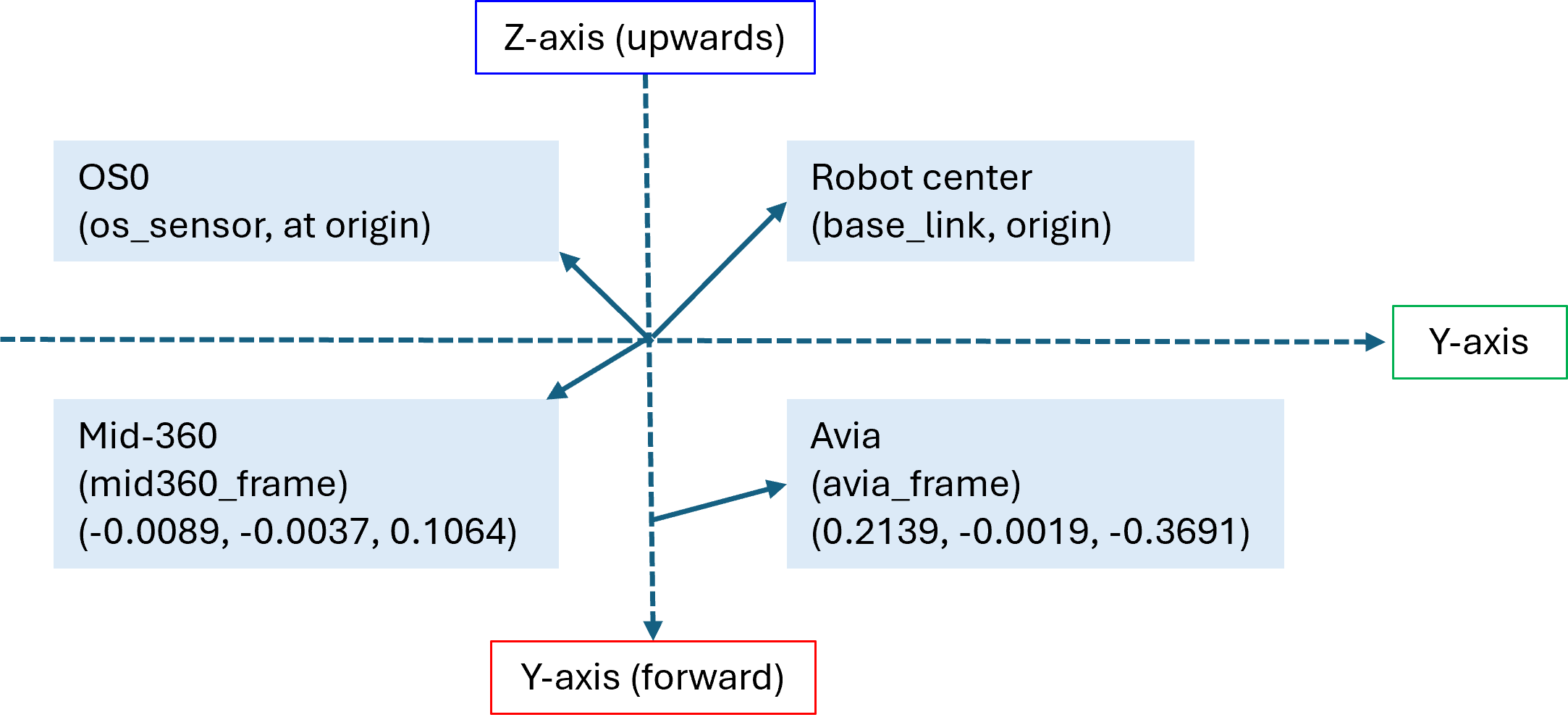}
    \caption{The calibrated data collection platform.}
    \label{fig:calibration}
\end{figure}

Table~\ref{tab:lidar-specification} shows the specification of the three distinct lidars. Apart from these main sensors, we also include the RealSense L515 camera, Xsens MTI-680G for IMU data, GNSS-RTK, and motion capture (mocap) data. 

\begin{table*}[t]
    \centering
    \caption{Specifications of the three Lidar sensors used in the study.}
    \label{tab:lidar-specification}
    \resizebox{0.92\textwidth}{!}{%
    \begin{tabular}{lccccccc}
        \toprule
        \textbf{Sensor} & \textbf{Type} & \textbf{FoV (H $\times$ V)} & \textbf{Max Range (m)} & \textbf{Power (W)} & \textbf{Supply Voltage (V)} & \textbf{Weight (g)} & \textbf{IMU Model} \\
        \midrule
        Ouster OS0-128 & Spinning & 360° $\times$ 90° (nominal) & Up to 240 & 14–20 (28 peak) & 10–51 & 430–500 & IAM-20680HT \\
        Livox Avia & Solid-state & 70.4° $\times$ 77.2° (non-repetitive) & Up to 450 & 8–9 (16 peak) & 10–15 (9–30 w/ conv.) & 498 & BMI088 \\
        Livox Mid-360 & Dome-shaped & 360° $\times$ (-7° to 52°) & Up to 70 & 6.5 & 9–27 & 265 & ICM40609 \\
        \bottomrule
    \end{tabular}
    }
    \vspace{-1.0em}
\end{table*}

\begin{figure*}[t]
    \centering
    \includegraphics[width=0.95\textwidth]{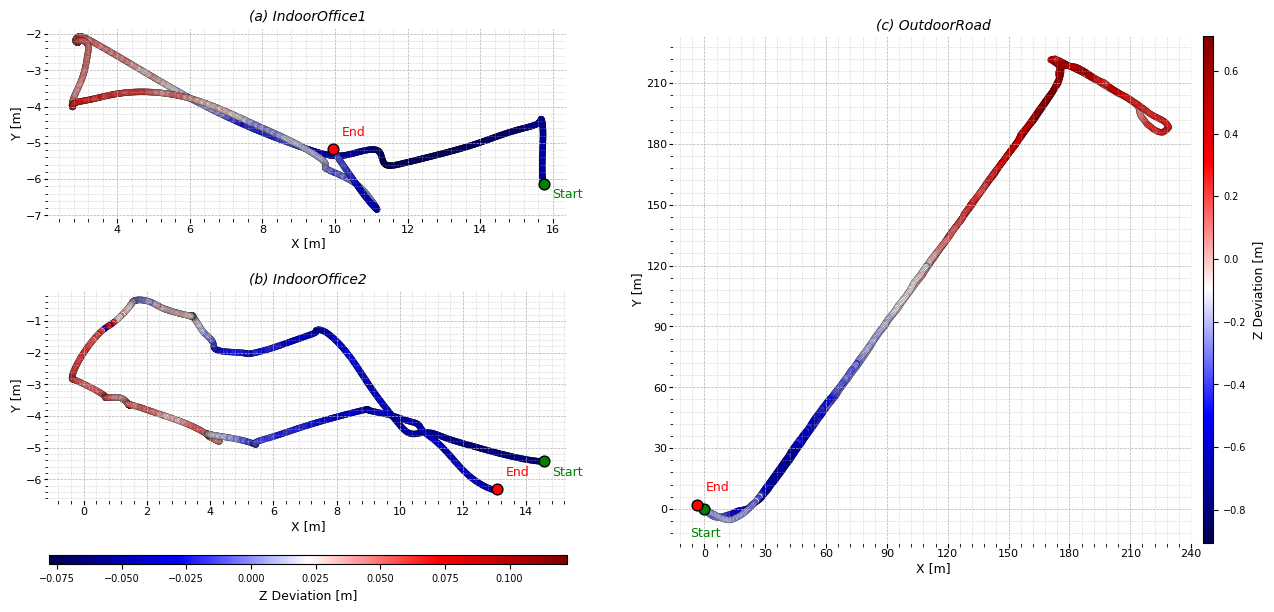}
    \caption{The designated ground truth paths of all the collected data sequences.}
    \label{fig:paths}
    \vspace{-1.5em}
\end{figure*}

\subsection{Software}
The software environment was built on Ubuntu 20.04 with ROS Noetic, which was the foundation for acquiring and processing sensor data. Several dependencies were integrated to enable efficient point cloud extraction, conversion, and trajectory evaluation. The EVO package~\footnote{https://github.com/MichaelGrupp/evo} was utilized to compute the Absolute Pose Error (APE), providing a standardized measure of odometry accuracy. Custom Python scripts were also developed to extract, format, and analyze the odometry outputs generated by various Iterative Closest Point (ICP) methods.

\subsection{Data Collection}
The sensors were rigidly mounted onto a customized sensor platform as shown in Fig.~\ref{fig:hardware}. The calibration results can be found in Fig.~\ref{fig:calibration}.
The timestamp synchronization among sensors is through Precision Time Protocol (PTP). 
Data collection was conducted across structured indoor and unstructured outdoor environments, specifically selected to challenge the lidar sensors across diverse conditions commonly encountered in robotic and mapping applications:
\subsubsection{Indoor Environment}
Indoor data collection took place within a controlled, factory-like experimental facility. A mocap system was deployed to provide accurate ground truth positional data, enabling high-precision evaluation and validation of lidar performance in structured scenarios.

\subsubsection{Outdoor Environment}
The outdoor data collection was performed along open-road settings characterized by mixed natural and urban features. Ground truth positioning data in this environment was acquired using the Xsens MTI-680G, which combines IMU and GNSS-RTK to provide high-frequency (100\, Hz) and high-precision (centimeter-level) reference trajectories suitable for evaluating lidar-based localization and SLAM algorithms under realistic conditions.



\subsection{SLAM Benchmarking and Evaluation Protocol}
To assess the odometry capabilities of different lidar types under realistic conditions, we benchmarked five state-of-the-art SLAM algorithms using our synchronized multi-lidar dataset. The dataset covers structured indoor environments (e.g., office settings) and unstructured outdoor scenes (e.g., open terrain), ensuring comprehensive evaluation across diverse operating scenarios encountered in robotics and autonomous systems.
We applied a uniform maximum lidar range of 60\, m for indoor data across all SLAM pipelines to simulate constrained sensing conditions typical of indoor environments.
For outdoor data, we used each sensor's default maximum sensing range (Livox Avia: 450\, m, Livox Mid-360: 100\, m, and Ouster OS0-128: 150\, m) to reflect their real-world deployment settings better.
These range constraints were enforced during preprocessing to ensure comparability while preserving each sensor’s practical sensing capabilities.

The SLAM methods evaluated span a range of architectural paradigms:

\begin{itemize}[leftmargin=*, align=left, itemsep=2pt, topsep=2pt]
\item \textbf{FAST-LIO2}\cite{xu2021fastlio2fastdirectlidarinertial}: A tightly coupled LiDAR-inertial odometry system using iterated Kalman filtering and surfel-based features.
\item \textbf{FASTER-LIO}\cite{9718203}: An efficiency-optimized version of FAST-LIO2 designed for faster computation.
\item \textbf{S-FAST-LIO}\cite{wang2023sfastlio}: A lightweight variant focused on real-time deployment in sparse environments.
\item \textbf{GLIM}\cite{KOIDE2024104750}: A hybrid SLAM approach that integrates global–local pose graph optimization with a LiDAR-based front end.
\item \textbf{FAST-LIO-SAM}~\cite{engcang2023fastliosam}: A fused architecture combining FAST-LIO’s front-end odometry with the loop closure and mapping backend of LIO-SAM.
\end{itemize}

\subsection{Point Cloud Matching Approaches}
This study evaluates three key ICP variants: Point-to-Point ICP, Point-to-Plane ICP, and Hybrid ICP, chosen for their complementary strengths in structured and unstructured environments. Their suitability under different geometric scenarios is well-established in prior research~\cite{121791,132043,SophiaZhangZhengyou1992}.

Point-to-Point ICP minimizes the Euclidean distance between corresponding points. Though efficient, it is sensitive to outliers and sparse data. We use the KISS-ICP package, which improves robustness through adaptive thresholding and a robust kernel function for outlier suppression~\cite{Vizzo_2023}. These enhancements make KISS-ICP particularly effective for real-time registration across diverse lidar modalities.

Point-to-Plane ICP improves alignment by incorporating local surface normals, which helps reduce geometric distortion. It is especially effective in environments rich in planar structures, such as corridors and urban settings~\cite{132043}. We implemented this using Open3D-GICP~\cite{Zhou2018,jelavic2022open3d}, which treats correspondences as Gaussian distributions for greater stability. Parameters such as correspondence distance and convergence thresholds were carefully tuned for each sensor.

Hybrid ICP, implemented using GenZ-ICP, blends point-to-point and point-to-plane strategies. It dynamically adjusts their weights based on scene characteristics, enhancing robustness in degenerate cases like narrow corridors~\cite{Lee_2025}. This flexibility makes it suitable for heterogeneous scan patterns, such as solid-state versus spinning LiDARs. We evaluated GenZ-ICP across varied indoor and outdoor scenes to test its adaptability.

To ensure fair comparison, all ICP methods underwent systematic hyperparameter tuning, including settings for correspondence distance, convergence criteria, and outlier rejection. The maximum correspondence range was fixed at 20\,cm for indoor data, which provided consistent accuracy across environments. For outdoor scenes, we uniformly applied a 60\,m range constraint across all ICP methods to simulate short-range local registration.

Because classical ICP lacks global trajectory optimization and loop closure, we limited its evaluation to short outdoor segments, specifically \textit{OutdoorRoad-cut0} and \textit{OutdoorRoad-cut1}, to avoid cumulative drift and reflect its intended use case. This strategy allowed us to assess local registration performance fairly, without conflating it with long-range tracking capabilities exclusive to SLAM systems. Our results support prior findings on the importance of tuning range limits to maintain ICP convergence stability.


\subsection{Evaluation Metrics}

Lidar-based odometry and point cloud matching accuracy were assessed by comparing the estimated trajectories to the ground-truth data. This ground-truth data was obtained using a motion capture system for indoor datasets and GNSS-RTK for outdoor environments. The primary metric used for this evaluation was the Absolute Pose Error (APE), which measures the deviation between the estimated and actual poses throughout the entire trajectory.
For each algorithm and lidar configuration, we report the mean and standard deviation of APE to provide insights into the accuracy and consistency of the pose estimates. 

%% file: sec/05_Experiments.tex
\input{tbs/tab-slams_indoor}

\section{Experimental Results}




\subsection{Dataset}~\footnote{The dataset will appear online soon.}
The dataset collected in this study is categorized into two primary subsets: indoor and outdoor environments. 
We collected two indoor sequences, \textit{IndoorOffice1} and \textit{IndoorOffice2}, featuring distinct trajectories in the same environment, and one outdoor sequence, \textit{OutdoorRoad}. All data were recorded using the ROS in rosbag format. 
The following outlines the data format for each sensor modality included in the dataset:

\textbf{Point Cloud Data:} Point cloud data format varies slightly across the different lidar sensors. For the Ouster OS0 (spinning lidar), each point includes the fields \texttt{x, y, z} (\texttt{float32}), \texttt{intensity}, \texttt{reflectivity}, \texttt{ring}, \texttt{ambient}, \texttt{range} and \texttt{t} (timestamp offset in nanoseconds). In contrast, the Livox Avia and Mid360 sensors (solid-state lidars) provide a point cloud consisting of \texttt{x, y, z, intensity}, and \texttt{tag, line} (\texttt{uint8}). Additionally, Mid360 provides a per-point \texttt{timestamp} (\texttt{float64}) in UNIX epoch nanoseconds.
    
\textbf{Inertial Data:} IMU data is available from all lidars using the \texttt{sensor\_msgs/Imu} message type. For Livox Avia and Mid360, their IMU data are published at 200\, Hz to  \texttt{/avia/livox/imu} and \texttt{/mid360/livox/imu}, respectively, while for Ouster OS0, the corresponding topic is \texttt{/ouster/imu}, published at 100\, Hz.
        

    
\textbf{Ground Truth:} Ground truth data is all in \texttt{geometry msgs::PoseStamped} format. They were obtained from a MoCap system for the indoor sequences, published under the topic \texttt{/vrpn_client_node/unitree_b1/pose}, and from a GNSS-RTK system for the outdoor sequence, available via \texttt{/gnss_pose}.

\input{tbs/tab-slams_outdoor}
\begin{figure}[h]
\centering
\begin{subfigure}[t]{0.24\textwidth}
\centering
    \def\plotwidth{\textwidth}
    \def\plotheight{1.4\textwidth}
    \input{tex/boxplot_slams}
\caption{SLAM (Indoor and Outdoor)}
\end{subfigure}
\hfill
\begin{subfigure}[t]{0.24\textwidth}
\centering
    \def\plotwidth{\textwidth}
    \def\plotheight{1.4\textwidth}
    \input{tex/boxplot_icps}
\caption{ICP (Indoor and Outdoor)}
\end{subfigure}
\caption{Grouped boxplots comparing LiDAR performance across indoor and outdoor environments for SLAM and ICP. Box colors represent Avia (red), Mid-360 (blue), and Ouster (green) sensors.}
\label{fig:lidar_vertical_filled}
\vspace{-1.0em}
\end{figure}
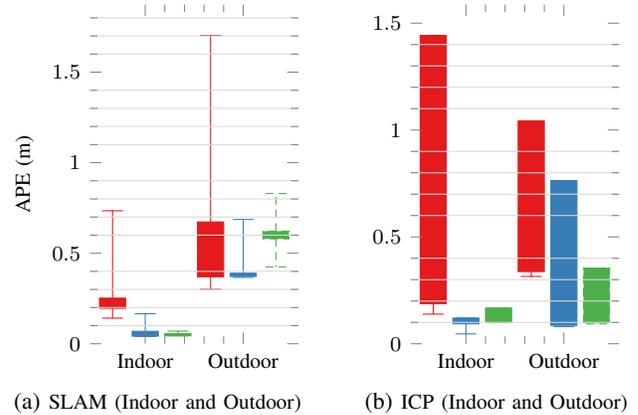

\subsection{SLAM Benchmarking}
\paragraph{Indoor SLAM Benchmark}

We evaluated five SLAM algorithms across three lidar types, Ouster OS0-128, Livox Avia, and Livox Mid-360, on the \textit{IndoorOffice1} and \textit{IndoorOffice2} datasets. The upper part of Table~\ref{tab:slam_combined_results} shows APE as mean $\pm$ std, with underlined values indicating the best overall and bold values the best per sensor.

Livox Mid-360 achieved the best overall APE performance, particularly under FAST-LIO2, FASTER-LIO, and S-FAST-LIO, due to its dome-shaped design and wide vertical FoV. This is confirmed by Fig.~\ref{fig:lidar_vertical_filled}(a), where Mid-360 (blue) shows the lowest median and tightest spread for indoor SLAM.

Ouster delivered stable per-method results with narrow distributions, seen in the green box plot. Avia, by contrast, suffered from larger error variances, particularly under GLIM, reflected in its wide red box with outliers. These trends highlight the robustness of dome-shaped sensors like Mid-360 in cluttered indoor environments.

\input{tbs/tab-icps_indoor}


\paragraph{Outdoor SLAM Benchmark}
The \textit{OutdoorRoad} dataset results are summarized in the lower part of Table~\ref{tab:slam_combined_results}.
Avia, benefiting from its 450\,m  range, achieves the best overall APE in one method, FASTER-LIO, where it outperforms both Mid-360 and Ouster, as reflected by the underlined and bold value. However, it performs significantly worse under S-FAST-LIO and GLIM, and shows higher variance in Fig.~\ref{fig:lidar_vertical_filled}(a), where the red box for Avia is tall and dispersed.
Mid-360 yields strong results in FASTER-LIO and S-FAST-LIO, and maintains relatively consistent performance across the board, though it does not dominate any single method overall. Its compact blue box plot suggests moderate accuracy with lower variance in outdoor settings.
Ouster achieves competitive results under FASTER-LIO, with both mean and std bolded, but it does not stand out from the other methods. Its wider green box in the figure indicates more variability. Overall, the results emphasize that while Avia can achieve high accuracy under specific conditions, Mid-360 remains more consistently reliable across algorithms.

\subsection{Point Cloud Matching Evaluation}
\paragraph{Indoor Point Cloud Matching Evaluation}

Three ICP pipelines, KISS-ICP, GenZ-ICP, and Open3D-GICP, were evaluated on IndoorOffice datasets. Table~\ref{tab:icp_indoor_results} reports APE, with underlines for best overall and bold for per-sensor best results.
Mid-360 ranks best overall, especially under KISS-ICP in indoor datasets, and GenZ-ICP in \textit{IndoorOffice1}. It also achieves competitive results under Open3D-GICP (Scan2Map), particularly in \textit{IndoorOffice2}. This is supported by the compact blue box in Fig.~\ref{fig:lidar_vertical_filled}(b), indicating low variance across indoor ICP tasks.

Ouster (green) performs well, especially in GenZ-ICP and Open3D-GICP on \textit{IndoorOffice1}, though with slightly more spread. Avia (red) performs the worst, particularly in Open3D-GICP (Scan2Scan), whose sparse vertical resolution results in high error and variance.






\paragraph{Outdoor Point Cloud Matching Evaluation}

We benchmarked the same ICP pipelines on \textit{OutdoorRoad-cut0} and \textit{OutdoorRoad-cut1} using a 60\,m range. Table~\ref{tab:icp_indoor_results} shows the APE statistics.
Mid-360 consistently ranks best or near-best across all methods. Notably, it achieves the lowest APE in Open3D-GICP (Scan2Map) on \textit{OutdoorRoad-cut1}, and leads in GenZ-ICP and KISS-ICP on OutdoorRoad-cut0. Its performance is confirmed by the blue box in Fig.~\ref{fig:lidar_vertical_filled}(b), reflecting tight error distributions in outdoor ICP.

Ouster performs best under Open3D-GICP (Scan2Map) in \textit{OutdoorRoad-cut1}, with the lowest mean among all sensors. It also shows strong stability under KISS-ICP in \textit{OutdoorRoad-cut0}, with a narrow error spread. However, its competitive edge is limited to specific configurations. Mid-360 remains the more consistently reliable performer across most ICP pipelines and segments. Avia again trails with higher and more variable error.
In contrast, Avia underperforms across all outdoor configurations, often showing the highest error and variability, especially under Open3D-GICP (Scan2Scan) due to its sparser scan structure and limited vertical coverage.

\begin{figure*}[ht]
    \centering
    \begin{subfigure}[b]{0.32\textwidth}
        \includegraphics[width=\textwidth]{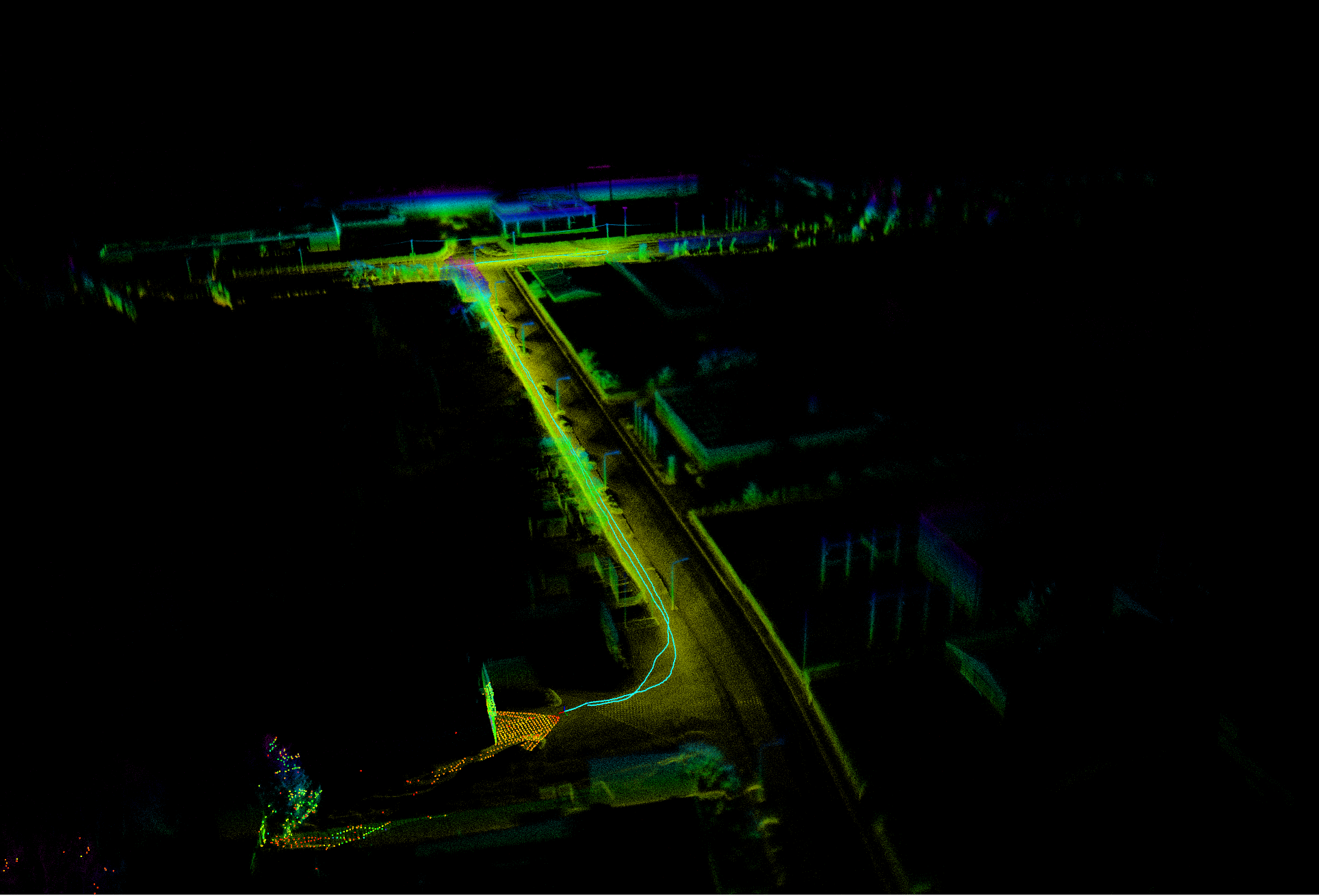}
        \caption{Avia (450\,m range)}
    \end{subfigure}
    \begin{subfigure}[b]{0.30\textwidth}
        \includegraphics[width=\textwidth]{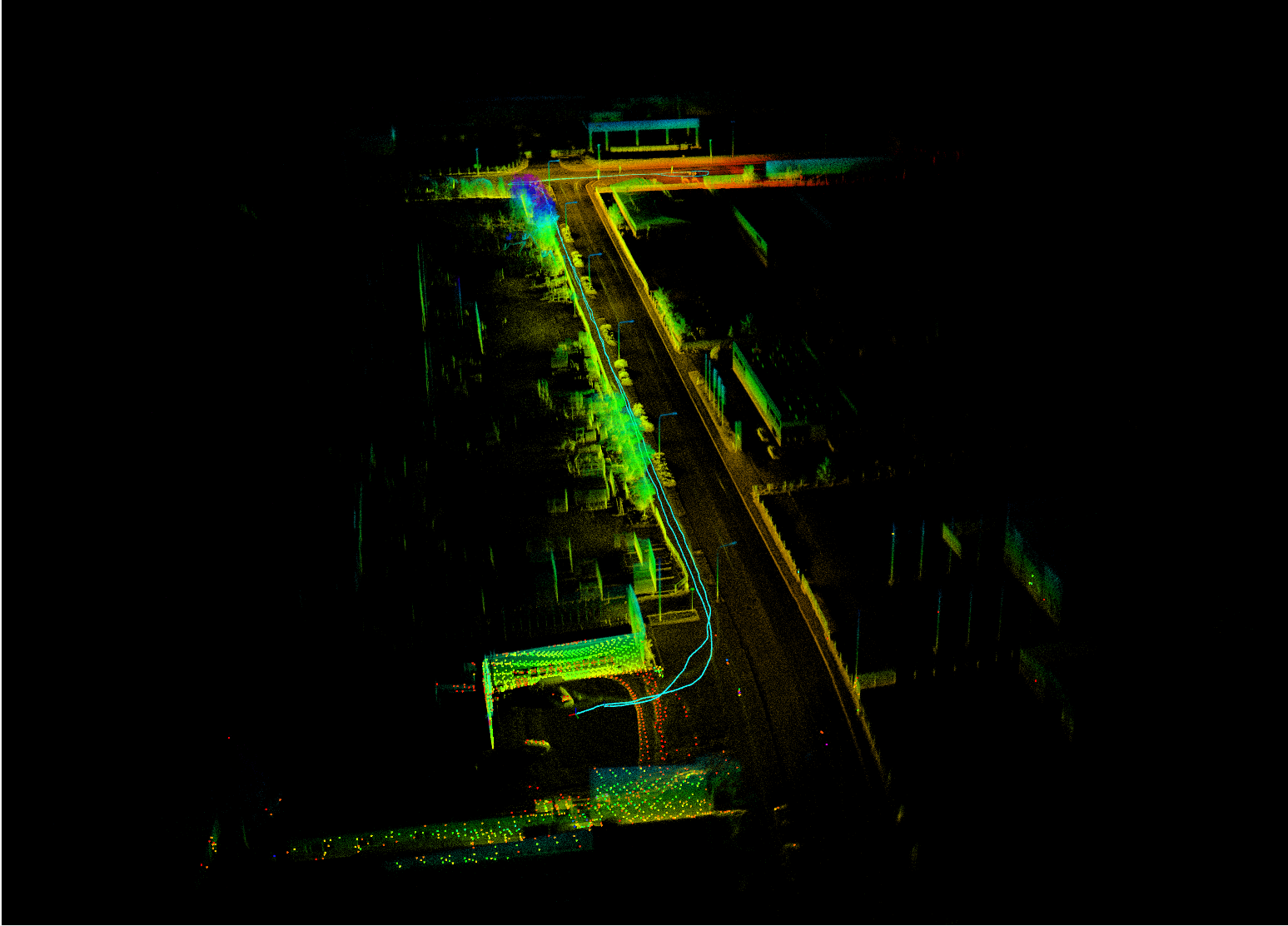}
        \caption{Mid-360 (100\,m range)}
    \end{subfigure}
    \begin{subfigure}[b]{0.31\textwidth}
        \includegraphics[width=\textwidth]{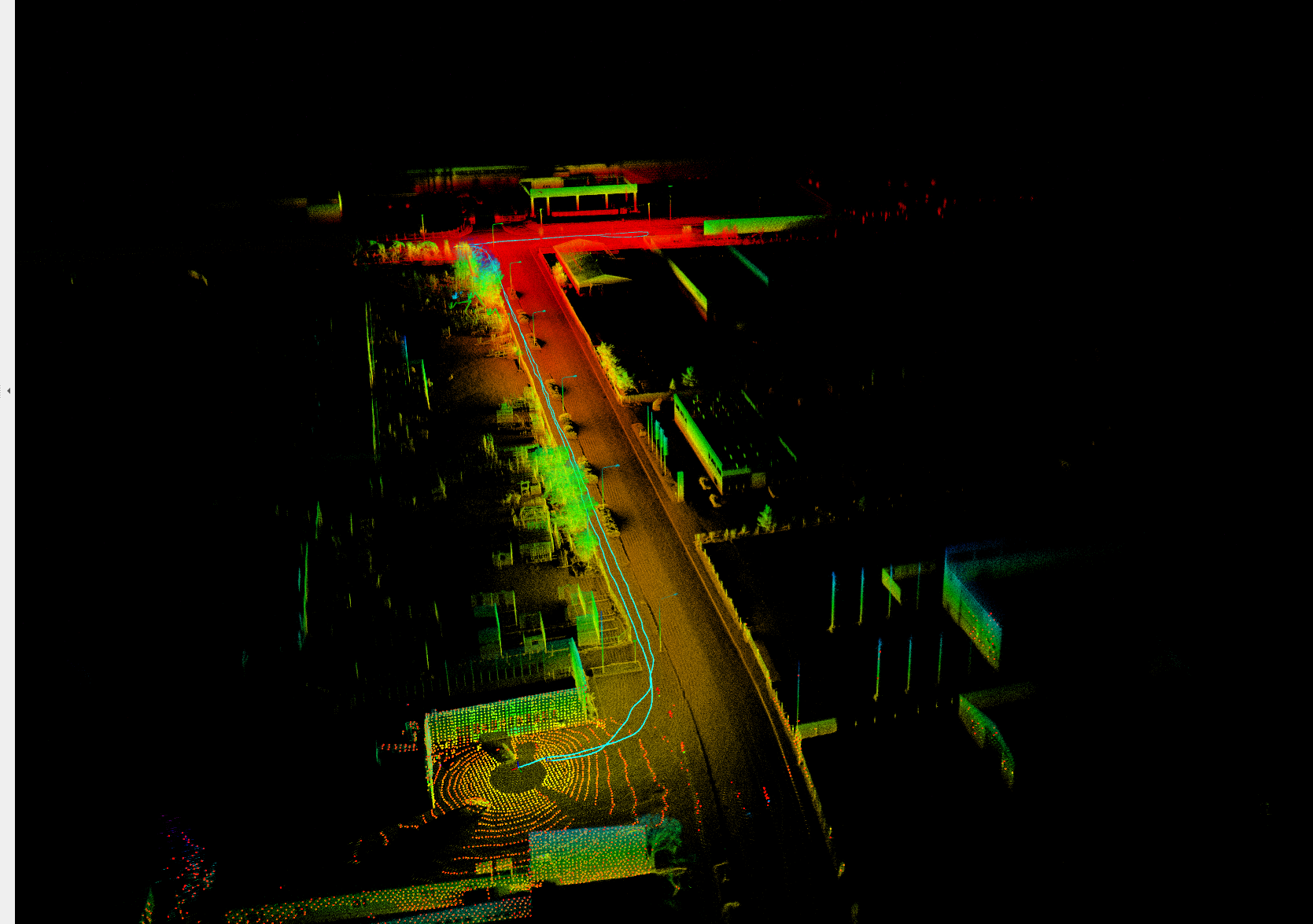}
        \caption{Ouster OS0-128 (150\,m range)}
    \end{subfigure}
    \caption{Outdoor SLAM map built by each lidar.}
    \label{fig:outdoor_cloud_examples}
\end{figure*}

%% file: tbs/tab-slams_indoor.tex
\begin{table*}[t]
    \centering
    \caption{APE (mean $\pm$ std) of lidar odometry in meters for various indoor and outdoor dataset settings. For each lidar type within each data sequence, the highest accuracy is indicated in bold. The overall best performance across different lidar types is both bolded and underlined.}
    \label{tab:slam_combined_results}
    \resizebox{\textwidth}{!}{
    \begin{tabular}{l|ccc|ccc}
        \toprule
        \multirow{2}{*}{\textbf{SLAM}}  
        & \multicolumn{3}{c|}{\textbf{\textit{IndoorOffice1}}} 
        & \multicolumn{3}{c}{\textbf{\textit{IndoorOffice2}}} \\
        & Avia\_10Hz & Mid360\_10Hz & Ouster\_10Hz 
        & Avia\_10Hz & Mid360\_10Hz & Ouster\_10Hz \\
        \midrule
        FAST\_LIO2       & 0.1436 $\pm$ 0.1390 & 0.0451 $\pm$ \textbf{\underline{0.0150}} & 0.0446 $\pm$ \textbf{0.0298} 
                        & 0.3581 $\pm$ 0.1334 & 0.0412 $\pm$ 0.0201 & 0.0438 $\pm$ 0.0471 \\
        FASTER\_LIO      & \textbf{0.1223} $\pm$ \textbf{0.0517} & 0.0918 $\pm$ 0.0423 & 0.0571 $\pm$ 0.0304 
                        & 0.2692 $\pm$ 0.1386 & 0.0460 $\pm$ \textbf{\underline{0.0147}} & 0.0549 $\pm$ 0.0491 \\
        S\_FAST\_LIO     & 0.1808 $\pm$ 0.1840 & \textbf{\underline{0.0427}} $\pm$ 0.0161 & 0.0541 $\pm$ 0.0367 
                        & \textbf{0.1019} $\pm$ \textbf{0.0536} & \textbf{\underline{0.0388}} $\pm$ 0.0190 & 0.0504 $\pm$ 0.0529 \\
        GLIM            & 0.3104 $\pm$ 0.2461 & 0.1760 $\pm$ 0.1734 & 0.0733 $\pm$ 0.0395 
                        & 1.1587 $\pm$ 0.6535 & 0.1559 $\pm$ 0.1142 & 0.0668 $\pm$ 0.0559 \\
        FAST\_LIO\_SAM  & 0.1436 $\pm$ 0.1392 & 0.0457 $\pm$ 0.0153 & \textbf{0.0442} $\pm$ 0.0299 
                        & 0.3581 $\pm$ 0.1327 & 0.0400 $\pm$ 0.0190 & \textbf{0.0431} $\pm$ \textbf{0.0398} \\
        \bottomrule
        \toprule
        \multirow{2}{*}{\textbf{SLAM}} & \multicolumn{6}{c}{\textbf{\textit{OutdoorRoad}}}\\
        & \multicolumn{2}{c}{Avia\_10Hz} & \multicolumn{2}{c}{Mid360\_10Hz} & \multicolumn{2}{c}{Ouster\_10Hz} \\
        \midrule
        FAST\_LIO2       
        & \multicolumn{2}{c}{0.3755 $\pm$ 0.1527} & \multicolumn{2}{c}{0.3893 $\pm$ 0.1788} & \multicolumn{2}{c}{0.5845 $\pm$ 0.3127} \\
        FASTER\_LIO      
        & \multicolumn{2}{c}{\textbf{\underline{0.3013}} $\pm$ \textbf{\underline{0.0818}}} & \multicolumn{2}{c}{\textbf{0.3666} $\pm$ 0.1668} & \multicolumn{2}{c}{\textbf{0.4245} $\pm$ \textbf{0.2273}} \\
        S\_FAST\_LIO     
        & \multicolumn{2}{c}{0.6730 $\pm$ 0.3205} & \multicolumn{2}{c}{0.3721 $\pm$ \textbf{0.1641}} & \multicolumn{2}{c}{0.6223 $\pm$ 0.3282} \\
        GLIM            
        & \multicolumn{2}{c}{1.7026 $\pm$ 0.2774} & \multicolumn{2}{c}{0.6867 $\pm$ 0.5828} & \multicolumn{2}{c}{0.8303 $\pm$ 0.4314} \\
        FAST\_LIO\_SAM  
        & \multicolumn{2}{c}{0.3700 $\pm$ 0.1545} & \multicolumn{2}{c}{0.3873 $\pm$ 0.1775} & \multicolumn{2}{c}{0.5795 $\pm$ 0.3137} \\
        \bottomrule
    \end{tabular}
    }
    \vspace{-1.0em}
\end{table*}

%% file: tex/boxplot_slams.tex
\definecolor{AviaColor}{RGB}{228,26,28}
\definecolor{Mid360Color}{RGB}{55,126,184}
\definecolor{OusterColor}{RGB}{77,175,74}

\begin{tikzpicture}
\begin{axis}[
    width=\plotwidth,
    height=\plotheight,
    ylabel={APE (m)},
    xtick={2,5},
    xticklabels={Indoor, Outdoor},
    xticklabel style={align=center},
    ymin=0,
    enlarge x limits=0.05,
    tick label style={font=\footnotesize},
    label style={font=\footnotesize},
    title style={font=\footnotesize},
    boxplot/draw direction=y,
    minor tick num=4,           
    minor grid style={gray!30}, 
    major grid style={gray!60}, 
    yminorgrids=true,
    axis on top,
    axis line style={draw=none},
]

\addplot+[boxplot prepared={
    lower whisker=0.1414,
    lower quartile=0.1958,
    median=0.2509,
    upper quartile=0.2509,
    upper whisker=0.7346},
    fill=AviaColor, draw=AviaColor] coordinates {} (axis cs:1,0);
\addplot+[boxplot prepared={
    lower whisker=0.0408,
    lower quartile=0.0429,
    median=0.0432,
    upper quartile=0.0689,
    upper whisker=0.1660},
    fill=Mid360Color, draw=Mid360Color] coordinates {} (axis cs:2,0);
\addplot+[boxplot prepared={
    lower whisker=0.0437,
    lower quartile=0.0442,
    median=0.0522,
    upper quartile=0.0560,
    upper whisker=0.0701},
    fill=OusterColor, draw=OusterColor] coordinates {} (axis cs:3,0);

\addplot+[boxplot prepared={
    lower whisker=0.3013,
    lower quartile=0.3700,
    median=0.3755,
    upper quartile=0.6730,
    upper whisker=1.7026},
    fill=AviaColor, draw=AviaColor] coordinates {} (axis cs:5,0);
\addplot+[boxplot prepared={
    lower whisker=0.3666,
    lower quartile=0.3721,
    median=0.3873,
    upper quartile=0.3893,
    upper whisker=0.6867},
    fill=Mid360Color, draw=Mid360Color] coordinates {} (axis cs:6,0);
\addplot+[boxplot prepared={
    lower whisker=0.4245,
    lower quartile=0.5795,
    median=0.5845,
    upper quartile=0.6223,
    upper whisker=0.8303},
    fill=OusterColor, draw=OusterColor] coordinates {} (axis cs:7,0);

\end{axis}

\end{tikzpicture}

%% file: tex/boxplot_icps.tex
\definecolor{AviaColor}{RGB}{228,26,28}
\definecolor{Mid360Color}{RGB}{55,126,184}
\definecolor{OusterColor}{RGB}{77,175,74}

\begin{tikzpicture}
\begin{axis}[
    width=\plotwidth,
    height=\plotheight,
    xtick={2,5},
    xticklabels={Indoor, Outdoor},
    xticklabel style={align=center},
    ymin=0,
    enlarge x limits=0.05,
    tick label style={font=\footnotesize},
    label style={font=\footnotesize},
    title style={font=\footnotesize},
    boxplot/draw direction=y,
    minor tick num=4,           
    minor grid style={gray!30}, 
    major grid style={gray!60}, 
    yminorgrids=true,
    axis on top,
    axis line style={draw=none},
]

\addplot+[boxplot prepared={
    lower whisker=0.1387,
    lower quartile=0.1880,
    median=0.4147,
    upper quartile=1.4423,
    upper whisker=1.4423},
    fill=AviaColor, draw=AviaColor] coordinates {} (axis cs:1,0);
\addplot+[boxplot prepared={
    lower whisker=0.0462,
    lower quartile=0.0934,
    median=0.1007,
    upper quartile=0.1201,
    upper whisker=0.1201},
    fill=Mid360Color, draw=Mid360Color] coordinates {} (axis cs:2,0);
\addplot+[boxplot prepared={
    lower whisker=0.0987,
    lower quartile=0.1019,
    median=0.1365,
    upper quartile=0.1672,
    upper whisker=0.1672},
    fill=OusterColor, draw=OusterColor] coordinates {} (axis cs:3,0);

\addplot+[boxplot prepared={
    lower whisker=0.3150,
    lower quartile=0.3378,
    median=0.7962,
    upper quartile=1.0423,
    upper whisker=1.0423},
    fill=AviaColor, draw=AviaColor] coordinates {} (axis cs:5,0);
\addplot+[boxplot prepared={
    lower whisker=0.0801,
    lower quartile=0.0863,
    median=0.0904,
    upper quartile=0.7634,
    upper whisker=0.7634},
    fill=Mid360Color, draw=Mid360Color] coordinates {} (axis cs:6,0);
\addplot+[boxplot prepared={
    lower whisker=0.0930,
    lower quartile=0.0984,
    median=0.0990,
    upper quartile=0.3545,
    upper whisker=0.3545},
    fill=OusterColor, draw=OusterColor] coordinates {} (axis cs:7,0);

\end{axis}
\end{tikzpicture}

%% file: tbs/tab-icps_indoor.tex
\begin{table*}[t]
    \centering
    \caption{Absolute Pose Error (APE) across ICP methods and lidars for indoor and outdoor data (mean $\pm$ std, in meters). For each lidar type within each data sequence, the highest accuracy is indicated in bold. The overall best performance across different lidar types is both bolded and underlined.}
    \label{tab:icp_indoor_results}
    \resizebox{\textwidth}{!}{
    \begin{tabular}{l|ccc|ccc}
    \toprule
    \multirow{2}{*}{\textbf{Method}}  & \multicolumn{3}{c|}{\textbf{\textit{IndoorOffice1}}} & \multicolumn{3}{c}{\textbf{\textit{IndoorOffice2}}}  \\
       &     Avia\_10Hz  &   Mid360\_10Hz     &    Ouster\_10Hz  &     Avia\_10Hz   &   Mid360\_10Hz     &   Ouster\_10Hz   \\
    \midrule
    KISS-ICP      & \textbf{0.1348} $\pm$ \textbf{0.1049}  & \textbf{\underline{0.0483}} $\pm$ \textbf{\underline{0.0405}} & 0.1042  $\pm$ 0.0838 & 0.6945  $\pm$ 0.2825 & \textbf{\underline{0.0441}} $\pm$ \textbf{\underline{0.0286}} & 0.0996 $\pm$ 0.0782 \\
    
    GENZ-ICP      & 0.2639 $\pm$ 0.1527 & 0.0653 $\pm$ 0.0588 & \textbf{0.0937}  $\pm$ \textbf{0.0511} & \textbf{0.1121} $\pm$ 0.0868 & 0.1214 $\pm$ 0.0586  & \textbf{0.1037 } $\pm$ \textbf{0.0764} \\
    
    Open3D-GICP (Scan2Map)      & 0.1387  $\pm$ 0.0738 & 0.1115 $\pm$ 0.0491  & 0.1082 $\pm$ 0.0613 & 0.1387 $\pm$ \textbf{0.0592} & 0.0899 $\pm$ 0.0425 & 0.1648  $\pm$ 0.0989 \\
    
    Open3D-GICP (Scan2Scan)      & 1.4341 $\pm$ 0.2804 & 0.1057 $\pm$ 0.0574  & 0.1161 $\pm$ 0.0524 & 1.4504 $\pm$ 0.9441 & 0.1345 $\pm$ 0.0787  & 0.2182 $\pm$ 0.1498 \\
    \midrule
        \multirow{2}{*}{}  & \multicolumn{3}{c|}{\textbf{\textit{OutdoorRoad-cut0}}} & \multicolumn{3}{c}{\textbf{\textit{OutdoorRoad-cut1}}}  \\
        \midrule
    KISS-ICP      & 0.3917 $\pm$ 0.3135  & \textbf{\underline{0.0545}} $\pm$ \textbf{\underline{0.0424}} & \textbf{0.0787} $\pm$ \textbf{0.0502} & \textbf{0.2840}  $\pm$ \textbf{0.2149} & 0.1058 $\pm$ 0.0628 & 0.1072 $\pm$ 0.0559 \\
    
    GENZ-ICP      & \textbf{0.1924} $\pm$ \textbf{0.1046} & 0.0645 $\pm$ 0.0489 & 0.0810  $\pm$ 0.1157 & 0.4376 $\pm$ 0.2479 & 0.1081 $\pm$ \textbf{\underline{0.0490}}  & 0.1157 $\pm$ 0.0603 \\
    
    Open3D-GICP (Scan2Map)      & 0.7787  $\pm$ 0.7435 & 0.0855 $\pm$ 0.0534  & 0.1177 $\pm$ 0.0655  & 0.8137 $\pm$ 0.4766 & \textbf{\underline{0.0953}} $\pm$ 0.0513 & \textbf{0.0803}  $\pm$ \textbf{0.0549} \\
    
    Open3D-GICP (Scan2Scan)      & 1.0489 $\pm$ 0.9901 & 1.0316 $\pm$ 0.3959  & 0.2015 $\pm$ 0.0900 & 1.0357 $\pm$ 0.8246 & 0.4952 $\pm$ 0.2269  &  0.5075 $\pm$ 0.2780 \\
    \bottomrule
    \end{tabular}
    }
    \vspace{-1.0em}
\end{table*}

%% file: sec/06_Conclusion.tex

\section{Conclusion}\label{sec:conclusion}
This study addresses the heterogeneity of lidar sensor types by introducing a novel dataset that, for the first time, includes data collected using three distinct categories of lidars: dome-shaped solid-state lidar, limitted FoV solid-state lidar, and spinning lidar. The dataset contains both indoor and outdoor environments with ground truth provided using MoCap in indoor settings and GNSS-RTK in ourdoor scenarios.
In addition to the dataset, we present a comprehensive SLAM benchmarking across all data sequences, utilizing multiple state-of-the-art SLAM algorithms. 
Results showed that the Livox Mid-360 consistently delivered high accuracy and stability across ICP and SLAM methods, particularly in complex scenes, owing to its wide vertical field of view and dense scan coverage. Livox Avia performed well in long-range SLAM scenarios but exhibited limitations in local registration. Ouster sensors achieved competitive results, especially under spinning-optimized SLAM pipelines, though their limited vertical resolution reduced performance in semi-structured environments.
Furthermore, to investigate the impact of lidar variability on geometric registration, we evaluate the point cloud matching performance of different lidars using several ICP (Iterative Closest Point) variants, without incorporating IMU assistance. Results suggest that the Mid-360 consistently achieves the most accurate alignment across a variety of test cases.

